\documentclass[10pt,twocolumn,letterpaper]{article}
\newcommand{\mycomment}[1]{}

\usepackage{cvpr}              
\usepackage{algorithm}
\usepackage{algorithmicx}
\usepackage{makecell}
\usepackage{bbding}
\usepackage{pifont}

\usepackage{algpseudocode} 
\algrenewcommand\algorithmicrequire{\textbf{Input:}}
\algrenewcommand\algorithmicensure{\textbf{Output:}}
\definecolor{cvprblue}{rgb}{0.21,0.49,0.74}
\usepackage[pagebackref,breaklinks,colorlinks,allcolors=cvprblue]{hyperref}

\def\paperID{19447} 
\def\confName{CVPR}
\def\confYear{2026}

\title{FAAR: Efficient Frequency-Aware Multi-Task Fine-Tuning via Automatic Rank Selection}

\author{Maxime Fontana\\
King's College London\\
London, UK \\
{\tt\small maxime.fontana@kcl.ac.uk}
\and
Michael Spratling \\
University of Luxembourg\\
Esch-sur-Alzette, Luxembourg\\
{\tt\small michael.spratling@uni.lu}
\and
Miaojing Shi\thanks{Corresponding author} \\
Tongji University \\
Shanghai, China \\
{\tt\small mshi@tongji.edu.cn}
}

\begin{document}
\maketitle
\begin{abstract}

Adapting models pre-trained on large-scale datasets is a proven way to reach strong performance quickly for downstream tasks. However, the growth of state-of-the-art models makes traditional full fine-tuning unsuitable and difficult, especially for multi-task learning (MTL) where cost scales with the number of tasks.
As a result, recent studies investigate parameter-efficient fine-tuning (PEFT) using low-rank adaptation to significantly reduce the number of trainable parameters. However, these existing methods use a single, fixed rank, which may not be optimal for different tasks or positions in the MTL architecture. Moreover, these methods fail to learn spatial information that captures inter-task relationships and helps to improve diverse task predictions. This paper introduces \textbf{F}requency-\textbf{A}ware and \textbf{A}utomatic \textbf{R}ank (\textbf{FAAR}) for efficient MTL fine-tuning. Our method introduces Performance-Driven Rank Shrinking (PDRS) to allocate the optimal rank per adapter location and per task. Moreover, by analyzing the image frequency spectrum, FAAR proposes a Task-Spectral Pyramidal Decoder (TS-PD) that injects input-specific context into spatial bias learning to better reflect cross-task relationships. Experiments performed on dense visual task benchmarks show the superiority of our method in terms of both accuracy and efficiency compared to other PEFT methods in MTL. FAAR reduces the number of parameters by up to 9 times compared to traditional MTL fine-tuning whilst improving overall performance. Our code is available \href{https://github.com/Klodivio355/FAAR}{here}.

\end{abstract}

\begin{figure}[!h]
  \centering
  \includegraphics[width=\linewidth]{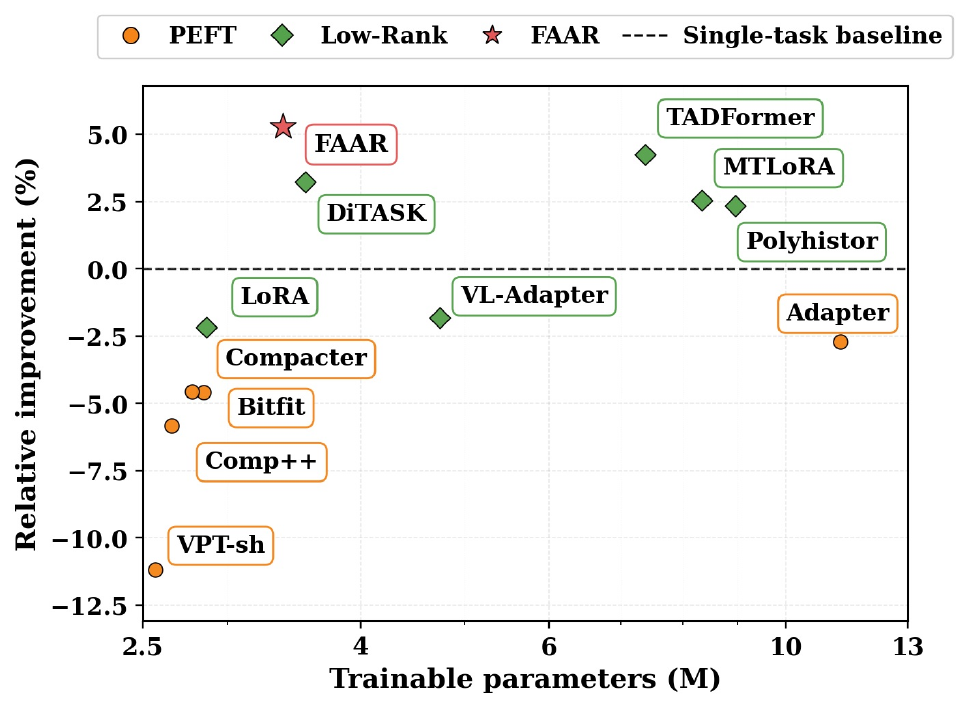}
  \caption{Accuracy/Efficiency trade-off comparison of FAAR with other PEFT MTL methods on PASCAL. FAAR achieves better overall performance thank traditional fine-tuning methods or MTL low-rank PEFT methods, while using fewer parameters.}
  \label{overview}
\end{figure}    
\section{Introduction}
\label{sec:intro}

Multi-Task Learning (MTL) aims to learn multiple tasks simultaneously. 
To achieve this, MTL seeks to partition representations into task-agnostic and task-specific features so that each task can utilize a common representation. This can help discover relationships and structures between tasks \cite{which_tasks, learning_to_branch}, which in return improves performance compared to task-specific models. MTL has been widely used to learn multiple dense visual tasks \cite{MTL-survey}, such as image segmentation \cite{image-seg-survey, semseg_survey}, depth estimation \cite{depth-survey} or surface normals estimation \cite{normals_paper}. Conventional models consist of a pretrained task-agnostic backbone, which is fine-tuned, attached with several task-specific decoders, trained from scratch.

Traditional fine-tuning involves weight updates for all backbone parameters. However, with the constant growth in the number of parameters in state-of-the-art backbones, this presents a significant challenge in terms of computational resources. 
To address this challenge, recent work has explored Parameter Efficient Fine-Tuning (PEFT), a paradigm that looks at updating a reduced proportion of backbone parameters for effective training~\cite{PEFT, PEFT2}. For example, some works have investigated prompt tuning, which consists of adapting pre-trained backbones by modifying the input with a set of learnable tokens \cite{prompt-tuning-survey}. Other works explore inserting adapters, which are small trainable networks, at various stages throughout backbones \cite{vl-adapter, adapter-is-all}. Low-Rank Adaptation (LoRA) \cite{lora-original} differs from traditional adapters by estimating the updates of the backbone weight matrices in a low-dimensional space via a simple matrix decomposition. This strategy comes from the observation that the updates to weight matrices made in traditional fine-tuning are inherently low-rank \cite{low_dim_justif, lora-original}. LoRA is the preferred choice due to its ease of implementation and performance/efficiency trade-off. Recently, a follow-up LoRA-inspired work, DoRA \cite{DoRA}, was proposed to decompose weights into directions and magnitudes in the fine-tuning optimization, leading to stable and strong performance, especially on low ranks. 

With the constant desire to increase the number of tasks to realize the robustness, efficiency and quality of MTL models, PEFT strategies have been explored in multi-task context to drastically reduce the significant cost of backbone fine-tuning \cite{efficientMTL, mtlora, adapter-vit, tadformer}. For example, VMT-adapter \cite{vmt-adapter} achieves this by using adapters. EfficientMTL \cite{efficientMTL} achieves this by learning to split model parameters and activations between tasks. Recent work has also expanded the LoRA paradigm to MTL \cite{mtlora,tadformer}.

Despite the promising and low-cost results obtained by previous LoRA-based MTL methods \cite{mtlora, tadformer}, there are limitations. First, these use a single rank for the entire fine-tuning process. This seems counter-intuitive from an accuracy perspective, as different tasks may require different adaptation strengths. Moreover, it also restricts each layer to have the same degree of flexibility, whereas one might expect different layers to require different amounts of fine-tuning. 
To cope with these issues, our method learns to reduce the rank for each task and layer dynamically by encouraging our fine-tuning to guide important adaptation components towards lower dimensions of the low-rank matrices. Our method prevails by learning to reduce the rank based on the overall performance of our MTL model. Therefore, it not only drastically reduces the number of trainable parameters as unnecessary adaptation is erased from the optimization scheme; but also improves performance on each task.To further improve performance on low ranks, we adapt DoRA \cite{DoRA} into the MTL context and build our method upon it. 

Second, we notice a lack of strong spatial inductive bias in previous LoRA-based MTL strategies \cite{mtlora, tadformer}, overlooking the power of harnessing cross-task interactions in deep layers~\cite{PAD-NET, MLoRE, MT-CP}.
To increase spatial awareness in vision models, Fast Fourier Transforms (FFTs) \cite{FFT} have been extensively used due to their ability to capture quality geometric information at low cost \cite{beta-FFT, FAD, FDConv}. Specifically, the frequency domain captures different scene information; while rapidly varying frequency values illustrate edges, smoother frequency variations demonstrate large patterns in the image. This information is particularly useful, as the former can be used to identify crisp boundaries for segmentation \cite{FDA}, or depth discontinuities for depth estimation \cite{Depth_FFT}. The latter, on the other hand, can be useful to capture both global scene layout and object locations \cite{FAR, wavelet}. Hence, we design a novel Task-Spectral Pyramidal Decoder, named TS-PD, to increase spatial bias within the adaptation at very low cost. First, TS-PD consists of a novel Channel-wise Spectral Filter (CW-SP) that harnesses powerful geometric cues for each task. We tailor our adapter to dense visual tasks by allowing each feature channel to learn the best frequencies to use. Second, we design a Cross-Task Consensus Alignment module (XT-Cons), a novel cross-task interaction that improves coherence between task predictions by learning to align task features on mutual frequencies. Our method is the first to leverage FFTs as part of dense visual MTL.

Our contributions are \textbf{threefold}: 
\begin{itemize}
  \item We introduce \textbf{F}requency-\textbf{A}ware and \textbf{A}utomatic \textbf{R}ank selection method (FAAR), a multi-task  PEFT method~\cite{DoRA}. FAAR first performs Performance-Driven Rank Shrinking (PDRS) which dynamically selects the smallest rank for each task and layer in a pre-trained backbone, while maximizing overall performance. 
  \item We design a novel Task-Spectral Pyramidal Decoder (TS-PD) which uses image frequency information to improve task-specific spatial requirements. In addition, TS-PD performs Cross-Task Consensus Alignment (XT-Cons), which allows tasks to align different task representations on mutual frequencies, improving coherence between task predictions.
  \item We demonstrate empirically that our method surpasses comparable multi-task baselines \cite{mtlora, tadformer} on two dense visual multi-task benchmark datasets both in terms of efficiency and accuracy as shown in \cref{overview}. 
\end{itemize}

\section{Related Work}
\label{sec:related-work}

\subsection{Parameter-Efficient Fine-Tuning in MTL}
\label{sec:peft-related}
MTL has been widely studied for dense vision tasks. 
Typical dense MTL models consist of a large pre-trained backbone of task-shared parameters. 
Previous work investigated PEFT strategies to reduce the major source of parameter budget, backbone fine-tuning. Firstly, adapters, which are small, trainable neural networks have been proposed to adjust MLP layers \cite{vmt-adapter} or modulate attention layers in a per-task manner \cite{adapter-vit}. Alternatively, EfficientMTL \cite{efficientMTL} suggests using adapters to adaptively select which activations and layers to split for each task \cite{efficientMTL}. Low-rank adaptation (LoRA) \cite{lora-original} offers a faster and lower-budget alternative. MTLoRA \cite{mtlora} uses LoRA in dense MTL and injects low-rank modules at diverse layer locations in a Swin Transformer \cite{swin-transformer}. TADFormer \cite{tadformer} expands on MTLoRA \cite{mtlora} by using dynamic convolutions \cite{dynamic-filter} to condition the matrix decomposition input on the tasks, thus alleviating the need for multi-task LoRA adapters introduced by MTLoRA \cite{mtlora}. However, these methods typically use a single, global rank which constrains the fine-tuning flexibility. Recently, DiTASK \cite{DiTASK} chooses not perform LoRA, but via a
few learnable parameters, performs singular value adaptation while preserving the pre-trained representations. On the other hand, our method preserves the predominant effectiveness of LoRA while minimizing the rank per-task and per-layer thereby preserving efficiency. 

\subsection{Automatic Rank Selection} 
Automatic Rank Selection aims to find the smallest rank for the LoRA strategies while maximizing performance. However, excessively reducing the rank can prevent adequate adaptation and lead to under-performance. Similarly, applying the same rank to all the adapted backbone layers poorly reflects the need for a different adaptation power throughout the backbone. Therefore, there has been research applied to single-task fine-tuning that aims to find optimal rank values. DyLoRA \cite{dylora} trains the backbone adaptation to be robust to smaller ranks by pushing important adaptation towards lower rank dimensions, allowing post-hoc rank choice without retraining. AdaLoRA \cite{AdaLoRA} sets ranks during training by pruning singular values \cite{SVD} according to an importance score, yielding layer-wise ranks. AutoLoRA \cite{autolora} attaches a selection variable to each rank component, optimizing these variables on validation loss to obtain per-layer ranks. Then, the matrices are retrained with the chosen ranks. While these methods aimed to improve single-task fine-tuning, our multi-task PEFT problem makes it even more important to find optimal rank values for each of the tasks. Our approach progressively shrinks the rank in a per-layer, per-task manner while using a performance signal during training to choose the best rank for our MTL objective.

\subsection{Fast Fourier Transforms}
\label{sec:frequency-related}
Recent research, in single-task learning, exploits FFTs due to their efficiency and capacity to capture complex geometric cues from an image \cite{macformer, FFR, FAD, FDConv}. High frequencies yield highly-detailed information about the image such as edges or contours, while low frequencies represent semantic, object-level information \cite{FDA, Depth_FFT}. FFT-based adapters have recently been proposed to take advantage of such information to guide adaptation. For example, FADA \cite{FADA} adapts vision models with explicit frequency adapters placed across layers to separate style features for domain-generalized segmentation. Adapt-ICMH \cite{adapt-icmh} proposes a plug-and-play spatial frequency adapter for downstream visual tasks, improving task-relevant spectral components. Finally, NightAdapter \cite{NightAdapter} inserts a frequency adapter that splits features, via discrete sine transform, into illumination-insensitive and sensitive bands and adapts them band-wise for night-time segmentation. While aligned with the aforementioned works, our method differs by being multi-task. This makes optimization more difficult, as different tasks need different frequencies. In addition, we propose aligning tasks using mutually beneficial frequency bands for better cross-task relationships.

\begin{figure*}[t]
  \centering

  \begin{subfigure}[t]{0.65\linewidth}
    \raggedright
    \includegraphics[
      width=\linewidth,
      height=.8\textheight,
      keepaspectratio,
      clip,trim=0 0 0 0
    ]{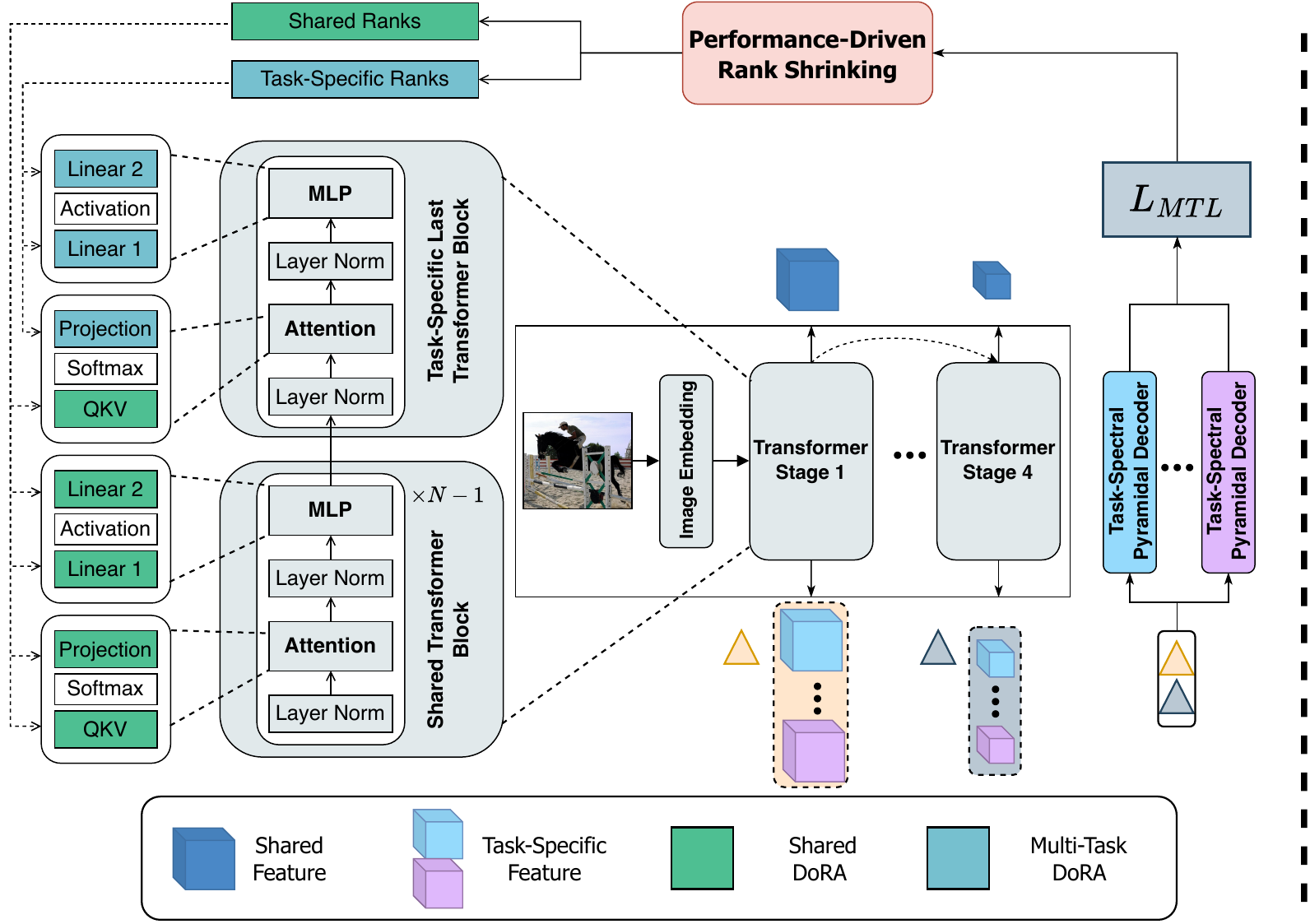}
    \caption{}
    \label{fig:overview_left}
  \end{subfigure}
  \hfill
  \begin{subfigure}[t]{0.32\linewidth}
    \raggedleft
    \includegraphics[
      width=\linewidth,
      height=.9\textheight,
      keepaspectratio,
      clip,trim=0 0 0 0
    ]{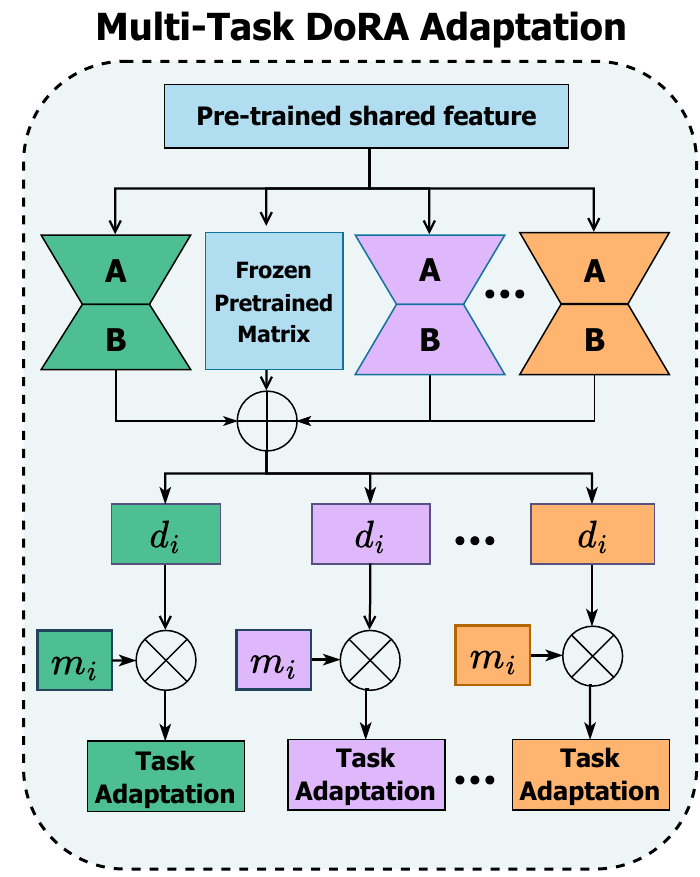}
    \caption{}
    \label{fig:overview_right}
  \end{subfigure}
  \caption{(a) Overview of our FAAR method. An input image embedding is given to a pre-trained backbone, in our case, a Swin Transformer \cite{swin-transformer}. At each stage of that backbone, representing a different resolution, DoRA \cite{DoRA} adapters are placed at attention and MLP layers (illustrated in \cref{fig:overview_right}). Up until the last block, shared feature adaptation takes place. The last block is kept for task-specific updates. Each shared and task-specific stage-wise representation are retained for our Task-Spectral Pyramidal Decoders (TS-PD). (b) Illustration of the DoRA adapters. For simplicity, only the multi-task version is shown. Shared adapters consist only of the \textcolor{ForestGreen}{green} path.}
  \label{fig:overview}
\end{figure*}

\section{Method}

\subsection{Overview}
The proposed method is illustrated in \cref{fig:overview} and consists of a frozen hierarchical ViT such as a Swin Transformer \cite{swin-transformer}. Each transformer stage consists of $N$ blocks, each of which is adapted. The last block of each stage is adapted in a per-task manner (illustrated in \cref{fig:overview_right}), while the earlier blocks use traditional DoRA adapters \cite{lora-original} to estimate weight updates that are shared across tasks. Each adapter, whether shared or task-specific, is associated with a learnable rank $r$ which controls the rank of the low-rank matrix decomposition. 
Following the backbone, a Task-Spectral Pyramidal Decoder (TS-PD) performs a frequency-based adaptation. To achieve this, a very low-cost frequency filter is attached to each of the task-specific, stage-wise representations which selects the best frequency bands for each of task to use. As part of this decoder, we introduce a Cross-Task Consensus Alignment method (XT-Cons), a FFT-based interaction module that allows tasks to share frequencies between them if they are useful to improve geometric consistency across task representations. Our whole fine-tuning strategy is controlled by our Performance-Driven Rank Shrinking (PDRS) method which consists in dynamically reducing the rank throughout training, based on the global objective loss $L_{MTL}$.

\mycomment{
\begin{figure}[t]
  \centering
  \begin{subfigure}[t]{0.80\linewidth}
    \centering
    \includegraphics[width=\linewidth]{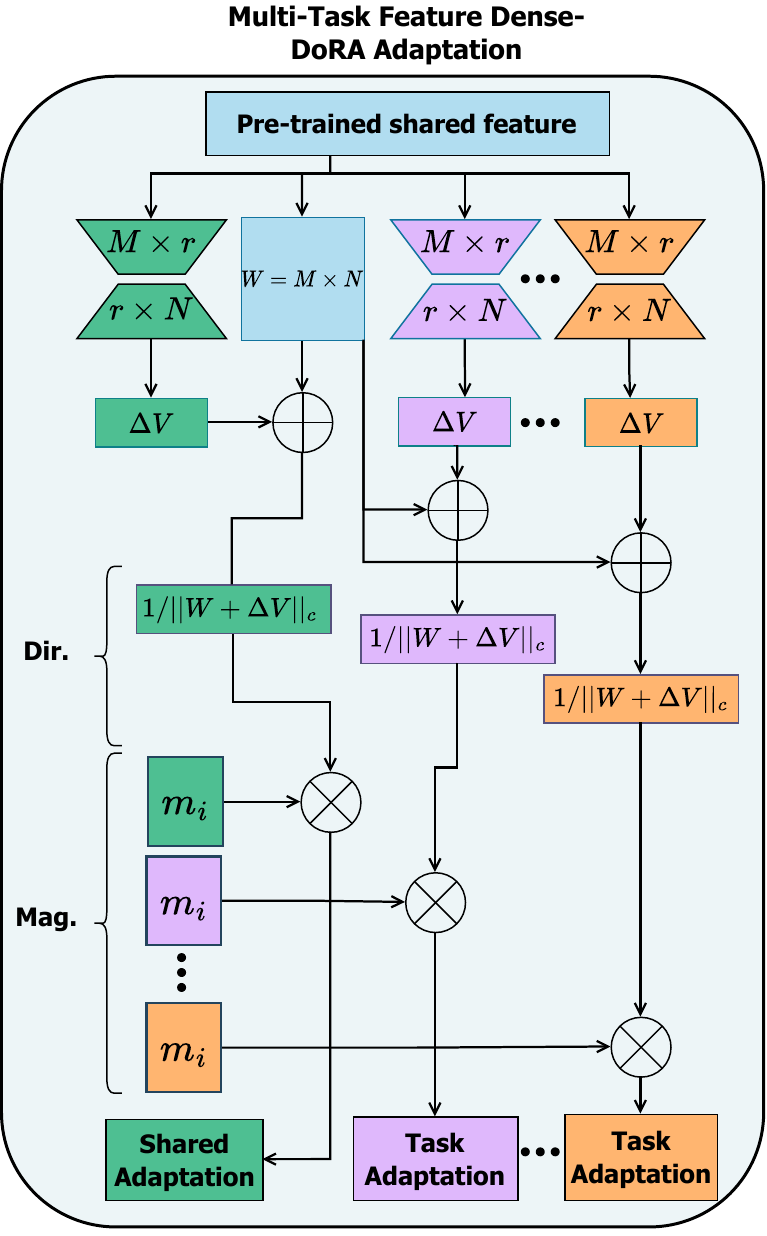}
    \caption{Multi-Task DoRA Adapter}
    \label{fig:lora_right}
  \end{subfigure}
  \caption{D$^{2}$oRA adapters adapt \textcolor{blue}{frozen}, pre-trained backbone networks. Following MTLoRA's design \cite{mtlora}, the shared adapter (on the left) is placed at each $N - 1$ blocks in a transfer stage. On the other hand, multi-task adapters (on the right) are placed at the last block of each transformer stage to retain task-specific representations. Note the $||\cdot||_{c}$ represents the column-wise norm of a matrix.}
  \label{fig:lora}
\end{figure}}

\subsection{Preliminary: Low-Rank Adaptation}
To efficiently adapt pre-trained features, we leverage the popular LoRA framework \cite{lora-original}. These modules are placed at attention and linear layers of transformer-based models, and calculate:
\begin{equation}
\begin{aligned}
Out^{LoRA}_{i} = W_{i}x + b_{i} + \alpha B_{i}A_{i}x,
\end{aligned}
\end{equation}
where the  pre-trained weight and bias matrices for layer $i$ are represented by $W_{i} \in  \mathbb{R}^{M \times N}$ and $b_{i} \in \mathbb{R}^{N}$, respectively. The adaptation residual is  calculated via two trainable matrices $A_{i}$ and $B_{i}$. $A_{i} \in \mathbb{R}^{M \times r}$ is a down-projection matrix that projects to a low-dimensional (rank) space. $B_{i} \in \mathbb{R}^{r \times N}$ is the up-projection matrix, which projects the low-rank estimate back to the frozen weight matrix space. However, this standard approach does not guarantee stability \cite{DoRA}, especially at very low ranks, which is crucial for our rank shrinking method described in \cref{sec:PDRS}. 
To address this issue DoRA \cite{DoRA} proposes to decouple magnitude from direction in the low-rank adaptation:

\begin{equation}
\begin{aligned}
\mathrm{Out}_{i}^{\mathrm{DoRA}} = m_{i}\,\frac{W_{i} + \alpha B_{i}A_{i}}{\bigl\|W_{i} + \alpha B_{i}A_{i}\bigr\|_2}\,x + b_{i},
\end{aligned}
\end{equation}
where $m_{i} \in \mathbb{R}^{N}$ defines the learned per-channel magnitude, $||\cdot||_{2}$ is the row-wise norm of a matrix, and the denominator is the direction term. We leverage DoRA as part of our PEFT method and embed it in both our shared and multi-task adapters as illustrated in \cref{fig:overview_right}.

\mycomment{
\begin{align}
\widehat{V}_{i} &= ||\!\bigl(||W_{i}||_{r} + \alpha B_{i}A_{i}\bigr)||_{r},\\
W_{i}^{DoRA} &= \operatorname{diag}(d_{i})\,\widehat{V}_{i},\\
\mathrm{Out}_{i} &= W^{\mathrm{DoRA}}_{i} x + b_{i},
\end{align}}

\mycomment{
\subsection{Dense-DoRA (D$^{2}$oRA)}
We observe that DoRA \cite{DoRA} is not suited for our experimental set-up. Firstly, the per-channel magnitude vector $d_{i}$ can take large ranges and therefore have no, or too large impact and therefore lead to destructive gradient behavior. However, our pre-trained backbone already exhibit some learnable patterns that we wish to preserve. Second, DoRA \cite{DoRA} scales the low rank effect depending on the norm of the original matrix. In other words, the low-rank update has a tiny effect on large norm rows and a big effect on small norm ones. We constrain both elements as shown in the equation below.
\begin{equation}
\begin{aligned}
\mathrm{Out}_{i}^{\mathrm{D}^{2}\mathrm{oRA}} 
&= \bigl(\|W_{i}\|_{2}\,d_{i}\bigr)\,
\frac{\,W_{i} / \|W_{i}\|_{2} + \alpha B_{i}A_{i}\,}
     {\bigl\|W_{i} / \|W_{i}\|_{2} + \alpha B_{i}A_{i}\bigr\|_{2}}\,
x + b_{i}.
\end{aligned}
\end{equation}
Here, we constrain $d_{i}$ on the norm of the input matrix $W_{i}$ to encourage the low-rank update to be scale invariant across rows. Similarly, we constrain $W_{i}$ so that the low-rank update has a more restrained update. As a result, we implement a more stable DoRa version which is more tailored for our dense visual task experimental set-up. We illustrate in \cref{fig:lora} how we leverage $D^{2}oRA$ as part of our adapters. Furthermore, we demonstrate in the following section how we manipulate $r$ as part of our rank shrinking method.}

\subsection{Performance-Driven Rank Shrinking (PDRS)}
\label{sec:PDRS}

\mycomment{
\begin{algorithm}[t]
\caption{FAAR: Automatic Rank Selection}
\label{alg}
\small
\begin{algorithmic}[1]
\Require minibatches $\mathcal{B}$; adapters $\mathcal{A}$
\Require ranks $R(a)$; targets $\rho(a)$; EMA $\beta$
\Require margin $m$; floor $f$
\State Init: for each $a\in\mathcal{A}$, cap $C(a)\leftarrow R(a)$, scores $q(a)\leftarrow 0$
\For{each step}
\State sample $(x,\cdot)\sim\mathcal{B}$
\For{each $a$}
\State $b \sim \mathrm{Unif}{1,,C(a)}$
\State $m=(\underbrace{1,\ldots,1}{b},0,\ldots,0)$
\State $A^=\mathrm{diag}(m),A_a$, \quad $B^=B_a,\mathrm{diag}(m)$
\State $\Delta W=B^A^$; forward; loss $\mathcal{L}$
\EndFor
\State backprop
\For{each $a$}
\For{$i=1$ to $b$}
\State $s^A_i=\big|\langle A{i,:},,\partial\mathcal{L}/\partial A_{i,:}\rangle\big|$
\State $s^B_i=\big|\langle B_{:,i},,\partial\mathcal{L}/\partial B_{:,i}\rangle\big|$
\State $s_i=\tfrac{1}{2},(s^A_i+s^B_i)$
\State $q_{a,i}\leftarrow \beta,q_{a,i}+(1-\beta),s_i$
\EndFor
\EndFor
\If{end of epoch}
\For{each $a$}
\State sort $q(a)$ desc; compute $c(k)=\frac{\sum_{j=1}^{k}q^\downarrow_j}{\sum_{j=1}^{R(a)}q^\downarrow_j}$
\State $K=\min{k:\ c(k)\ge \rho(a)}$
\State $k=\max(f,\ \min(C(a),\ K+m))$
\State $C(a)\leftarrow k$
\EndFor
\EndIf
\EndFor
\end{algorithmic}
\end{algorithm}
}

Recent MTL PEFT strategies inspired by LoRA \cite{mtlora, tadformer} apply the same rank $r$ to every layer. This is likely to be sub-optimal as previous work \cite{attention_in_trans, spot_tune} highlights the superiority of deeper layers for adaptation. It implies that all the tasks must be adapted at the same strength. However, this assumption goes against previous observations that some tasks are generally of different difficulty in MTL frameworks \cite{DTP, ABTPB, AdamV}. And as a result, we propose to dynamically reduce an initial manually-set rank, based on the MTL performance to \textit{(i)} drastically reduce the computational cost of the adaptation, \textit{(ii)} optimize the fine-tuning in a per-task and per-layer fashion. 
\cref{sec:rank-masking} describes how we effectively encourage our model to learn important adaptation power toward lower ranks of the  matrix decomposition. \cref{sec:coverage} describes the dynamic decision making process to decrease the rank.

\subsubsection{Rank Masking}
\label{sec:rank-masking}
In a low-rank matrix decomposition, the corresponding column $a \in \mathbb{R}^{r}$ in $A$ and row $b \in \mathbb{R}^{r}$ in $B$, define a rank-1 update: $a \otimes b$. Here, we describe how we progressively mask out rank-1 updates during training to reduce the computational cost and allow for flexible fine-tuning. Let $r_{curr}$ define a current rank value, at any time, during fine-tuning. We construct, at every forward pass, a binary masking vector representing active ranks for a given batch. To achieve this, we randomly sample a prefix size $b \in {1,\ldots,r_{\mathrm{curr}}}$, and form a binary prefix mask $m\in\{0,1\}^{r_{init}}$ with the first $b$ entries set to $1$. We then use this mask to apply rank-1 masking so that only the active prefix contributes to the residual. This mask strategy can be mathematically formulated as:
\begin{equation}
\begin{aligned}
A^{\mathrm{eff}} = \mathrm{diag}(m)A,\qquad
B^{\mathrm{eff}} = B\mathrm{diag}(m)
\end{aligned}
\end{equation}
\mycomment{
As a result, the resulting efficient masked low-rank update can be summarized as follows.
\begin{align}
\Delta W_i &= B^{\mathrm{eff}}_i\,D^{2}\,A^{\mathrm{eff}}_i, \\
\mathrm{Out}_i &= W_i\,x + b_i + \alpha B^{\mathrm{eff}}_i\,D^{2}\,A^{\mathrm{eff}}_i\,x, \\
\text{s.t.}\quad D &= \operatorname{diag}(m)\in\mathbb{R}^{r\times r}.
\end{align}}
This masking strategy allows the learning of important rank-1 updates to be pushed towards lower ranks. This encourages our fine-tuning to constraint its budget while still maintaining robustness to low ranks. We explain how we embed this rank masking strategy into our rank shrinking decision making method in the next section.

\subsubsection{Coverage Strategy}
\label{sec:coverage}
There is a need to meaningfully reduce the ranks in every layer, as naively cropping $A$ and $B$ after every forward pass is very noisy and not demonstrative of the performance of the model. Therefore, to address this issue, after each back-propagation stage, we compute a per-rank update importance slot for each of the active ranks $i \le b$ using a first-order directional measure as detailed below:
\begin{equation}
\begin{aligned}
s_i = \tfrac{1}{2}\left(
\big\lvert \left\langle A^{\mathrm{eff}}_{:,i},\, \tfrac{\partial \mathcal{L}}{\partial A^\mathrm{eff}_{:,i}} \right\rangle \big\rvert
+
\big\lvert \left\langle B^{\mathrm{eff}}_{i,:},\, \tfrac{\partial \mathcal{L}}{\partial B^{\mathrm{eff}}_{i,:}} \right\rangle \big\rvert
\right).
\end{aligned}
\end{equation}
Here, directional derivatives represent local sensitivities for each of the rank-1 updates in (a column of) $A$ and (a row of) $B$ with respect to MTL loss. In other words, if each term has a large magnitude, it means that changing this row or column $i$ significantly changes the loss. Therefore, $s_{i}$ represents the loss-sensitivity for a rank-1 update $i$. To avoid aggregating noisy, per-forward loss-sensitivity values, we use an exponential moving average (EMA) to accumulate values across batches as defined below.
\begin{equation}
\begin{aligned}
\hat{s}_i \leftarrow \beta \hat{s}_{i-1} + (1 - \beta) s_i
\end{aligned}
\end{equation}
In this equation, $\beta \in [0,1)$ represents the decay rate. A larger $\beta$ represents a smoother average.
At the end of each epoch, we sort the EMA scores in descending order to put the most important rank-1 updates first. Subsequently, we perform top-k selection over this representation, as follows: 
\begin{equation}
\begin{aligned}
\label{rho_eq}
c(k) = \frac{\sum_{j=1}^{k} \hat s_{(j)}}{\sum_{j=1}^{r} \hat s_{(j)}}, \qquad
K = \min \{\, k \,:\, c(k) \ge \rho \,\}.
\end{aligned}
\end{equation}
Here, we manually define the coverage criterion $\rho$ and select the minimum number of $K$ slots that satisfy the coverage criterion on the left. In other words, we keep a minimum amount of rank-1 updates that preserve the important loss changes. As these loss changes correlate with performance, they present a meaningful shrinkage criterion. Finally, computational cost is drastically reduced as uncovered ranks are permanently erased from the optimization scheme. As a result, for the next epoch, we set $r_{curr} = k$ and so on. Such strategy trains our MTL model to be robust by regularly exposing the model to smaller active ranks. Each DoRA adapter, whether task-specific or shared, can leverage the best rank that improves the overall objective. Therefore, this results in a meaningful, performance-aware automatic rank selection.

\subsection{Task-Spectral Pyramidal Decoder (TS-PD)}
\begin{figure}[!t]
\centering
\includegraphics[width=\linewidth]{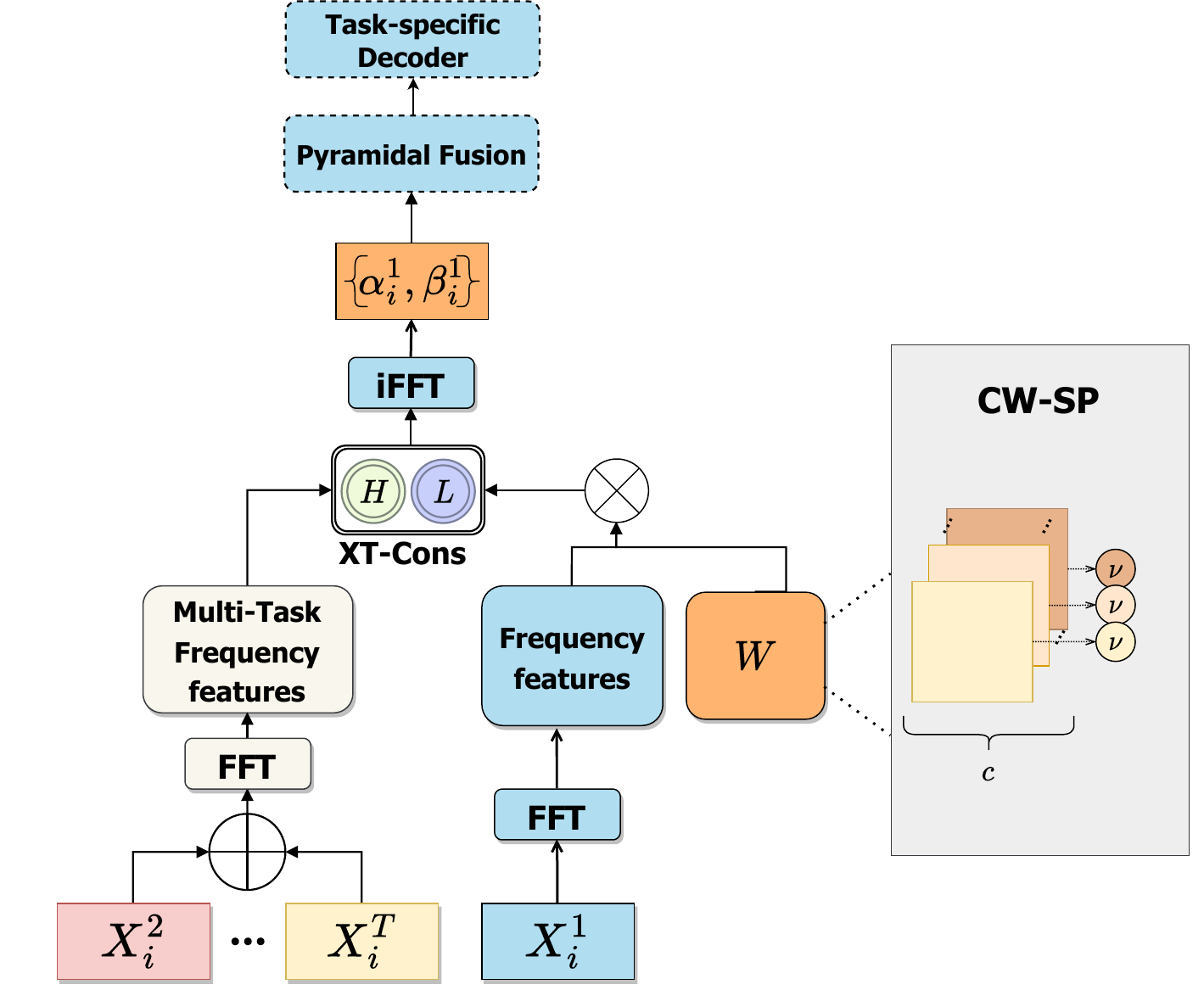}
\caption{Our Task-Spectral Pyramidal Decoder (TS-PD) consists of a Channel-Wise task-specific Spectral Filter (CW-SP) and a Cross-Task Consensus Alignment (XT-Cons) module to maintain geometric consistency.}
\label{FFT}
\end{figure}
We identify that previous works which aim to keep the fine-tuning low-cost, struggle to model strong spatial awareness. For the same reasons, previous works do not leverage cross-task relationships adequately for good MTL performance \cite{PAD-NET, MLoRE, taskExpert}. Therefore, inspired by the ability of frequency-based filters to capture good geometric cues at very-low cost \cite{FAD, FDConv}, we propose to leverage FFTs as a cheap way to address both issues. \cref{sec:TS_FT} introduces a novel frequency-based adapter, named Channel-wise Task-specific Spectral Filter (CW-SP), which allows fine spectral granularity, making it specifically designed for each of the dense visual tasks in our model. Then, to maintain cross-task consistency in the geometric space, \cref{sec:XTSRS} introduces a cross-task spectral alignment module which learns to align task representations on mutually beneficial frequencies.

\subsubsection{Channel-wise Spectral Filter (CW-SP)}
\label{sec:TS_FT}
In order to increase the spatial awareness in our model at a cheap cost, we leverage a decoder-based spectral filter which operates channelwise for better adaptation to dense visual tasks. To spatially adapt a task-specific feature, we design a frequency filter, as illustrated illustrated in \cref{FFT}. To tailor it to dense tasks, our frequency filter is learning 2D frequency filter per channel. Conceptually this allows for different feature maps of an image to leverage different frequencies. Specifically, given an image $I \in \mathbb{R}^{H \times W \times C}$, we compute the discrete Fourier Transform $FFT(I)$. Once the frequency domain obtained, we train a task-specific, resolution-specific matrix $W^{res}_{t}$. This matrix learns to scale different parts of the frequency spectrum for different frequency channels for the input $I$ via element-wise multiplication $Y= W\odot FFT(I)$. We then project the resulting representation back into the feature space by performing the inverse Fourier transform $Z = iFFT(Y)$. Finally, in order to modulate the importance of our channel-wise filter, we learn a scale and shift parameter per channel, $\alpha^{res}_{t}$ and $\beta^{res}_{t}$ respectively, and apply it such as : $Out^{res}_{t} = Z \times \alpha^{res}_{t} + \beta^{res}_{t}$.

\subsection{Cross-Task Consensus Alignment (XT-Cons)}
\label{sec:XTSRS}
We aim to maintain geometric consistency across tasks. To achieve this, we embed a frequency-based cross-task mechanism inside our TS-PD module. For each TS-PD, we leverage other task information as auxiliary information to get diverse scene information. For any main task representation ($X_{i}^{1}$ at resolution $i$), we get other task representations for the same resolution $X_{i}^{2}...X_{i}^{T}$. Subsequently, we align the main task based on the high and low frequency consensus of the auxiliary tasks, that is, the frequencies that auxiliary tasks agree on. To achieve this, we compute $F_{avg}$, the average representation of the auxiliary frequency spectrums. This gives us a general observation of the other task frequencies. Subsequently, we compute two masks from our main representation frequency spectrum: $M_{low}$ and $M_{high}$. Each of these are computed as thresholded representation of the frequency spectrum \textit{i.e.} $M_{low} = FFT(X_{i}^{1} )>0.5$:
\begin{equation}
\begin{aligned}
\Delta_{low, high} = M_{low, high} * (F_{avg} - FFT(X_{i}^{main})
\end{aligned}
\end{equation}
This process pushes the main representation toward what other tasks agree in the frequency domain on two basis, low and high frequencies. Subsequently, after projecting back $\Delta_{low, high}$ on to feature space, we scale their contributions with learnable scalars $\alpha_{low, high}$. Our final representation is therefore for a task $t$:
\begin{equation}
\begin{aligned}
Out^{res}_{t} = Out^{res}_{t} + \alpha_{low, high} \Delta_{low, high}.
\end{aligned}
\end{equation}
Resulting operations are then used for traditional pyramidal fusion and convolution-based decoding. As a result, our TS-PD is a powerful, yet cheap, dense spectral filter leveraging cross-task relationships for effective MTL. 

\begin{table*}[t]
\centering
\caption{Performance comparison on PASCAL-Context \cite{pascal-context} using \textit{Swin-Tiny} pretrained on ImageNet-1k as the backbone. We compare all methods with respect to $\Delta m$, the relative improvement vs.\ single-task full fine-tuning. FAAR improves all tasks with fewer parameters.}
\label{tab:multitask_pascal}
{\setlength{\tabcolsep}{3.6pt}
\renewcommand{\arraystretch}{1.0}
\small
\begin{tabular}{c|cccc|c|c|c}
\hline
\noalign{\vskip 0pt} 
\rule{0pt}{2.2ex}\textbf{Method} &
\shortstack{\textbf{SemSeg}\\\textit{(mIoU $\uparrow$)}} &
\shortstack{\textbf{HumanParts}\\\textit{(mIoU $\uparrow$)}} &
\shortstack{\textbf{Saliency}\\\textit{(mIoU $\uparrow$)}} &
\shortstack{\textbf{Normals}\\\textit{(rmse $\downarrow$)}} &
\shortstack{$\Delta m$\\(\%)} &
\shortstack{\textbf{Trainable}\\\textbf{Parameters (M)}} &
\shortstack{\textbf{Single inf.}\\\textbf{(all tasks)}} \\
\hline
Single Task & 67.21 & 61.93 & 62.35 & 17.97 & 0 & 112.62 & \ding{53} \\
MTL - Tuning Decoders Only & 65.09 & 53.48 & 57.46 & 20.69 & -9.95 & 1.94 & \checkmark \\
MTL - Full Fine Tuning      & 67.56 & 60.24 & 65.21 & 16.64 & +2.23 & 30.06 & \checkmark \\
\hline
Adapter \cite{adapter}          & 69.21 & 57.38 & 61.28 & 18.83 & -2.71 & 11.24 & \ding{53} \\
Bitfit \cite{bitfit}            & 68.57 & 55.99 & 60.64 & 19.42 & -4.60 & 2.85  & \ding{53} \\
VPT-shallow \cite{VPT}          & 62.96 & 52.27 & 58.31 & 20.90 & -11.18 & 2.57 & \ding{53} \\
VPT-deep \cite{VPT}             & 64.35 & 52.54 & 58.15 & 21.07 & -10.85 & 3.43 & \ding{53} \\
Compacter \cite{compacter}      & 67.26 & 56.41 & 60.08 & 19.22 & -4.55 & 2.78 & \ding{53} \\
Compacter++ \cite{compacter}    & 68.08 & 55.69 & 59.47 & 19.54 & -5.84 & 2.66 & \ding{53} \\
LoRA \cite{lora-original} ($r{=}4$) & 70.12 & 57.73 & 61.90 & 18.96 & -2.17 & 2.87 & \ding{53} \\
VL-Adapter \cite{vl-adapter}    & 70.21 & 59.15 & 62.29 & 19.26 & -1.83 & 4.74 & \ding{53} \\
HyperFormer \cite{Hypernetwork} & 71.43 & 60.73 & 65.54 & 17.77 & +2.64 & 72.77 & \ding{53} \\
Polyhistor \cite{polyhistor}    & 70.87 & 59.54 & 65.47 & 17.47 & +2.34 & 8.96 & \ding{53} \\
\hline
MTLoRA \cite{mtlora} ($r{=}64$)  & 67.90 & 59.84 & 65.40 & 16.60 & +2.55 & 8.34 & \checkmark \\
TADFormer \cite{tadformer} ($r{=}64$) & 70.82 & 60.45 & 65.88 & 16.48 & +4.24 & 7.38 & \checkmark \\
DiTASK \cite{DiTASK} ($r{=}64$) & 69.66 & \textbf{62.02} & 65.00 & 17.10 & +3.22 & 3.55 & \checkmark \\
\hline
\textbf{FAAR} ($r_{init}=64$) & \textbf{72.02} & 61.25 & \textbf{66.11} & \textbf{16.35} & \textbf{+5.28} & \textbf{3.38} & \checkmark \\
\hline
\end{tabular}
}
\end{table*}

\section{Experiments}
\label{sec:experiments}
\subsection{Implementation Details}
\label{sec:implementation_details}
\textbf{Datasets}. 
We evaluate FAAR on two widely-used MTL datasets. Firstly, PASCAL-Context \cite{pascal-context}. This dataset includes the following 4 tasks: semantic segmentation, human part segmentation, saliency distillation and surface normal estimation. It contains 4998 training images and 5105 for validation. Secondly, NYUDv2 dataset \cite{nyud}. This dataset covers semantic segmentation, surface normals estimation and monocular depth estimation. It has 795 training images and 654 validation images.

\noindent \textbf{Evaluation metrics}.
Following conventional MTL evaluation \cite{MTL-survey}, we use the mean Intersection over Union (mIoU) for semantic segmentation, saliency estimation and human part segmentation tasks. For surface normals estimation and depth estimation, we use the root mean squared error (rmse). To measure the overall performance of our model, we calculate the average per-task reduction in performance, noted $\Delta m$ relative to the single-task baseline, noted $st$. We detail this metric as follows:
\begin{equation}
\begin{aligned}
\Delta m = \frac{1}{T} \sum_{i=1}^{T} (-1)^{l_{i}} (M_{i} - M_{st, i}) / M_{st, i}
\end{aligned}
\end{equation}
where $l_{i} = 1$ for tasks for which a lower metric value means better performance and 0 otherwise as described in \cite{mtlora}.

\noindent \textbf{Implementation}.
\label{sec:imp}
We use a \textit{Tiny}-Swin Transformer \cite{swin-transformer}, pre-trained on the ImageNet dataset \cite{imagenet}, as a backbone and we use HRNet \cite{hrnet} decoders for each of the task, following other PEFT MTL works \cite{mtlora, tadformer, DiTASK}. All experiments were carried out on a single NVIDIA A40. We use a learning rate of $5e^{-4}$. Our batch size is 32. Our model contains a few hyper-parameters, $\rho_{task}$ and $\rho_{shared}$, as defined in \cref{rho_eq} which represent the coverage criterion for task and shared layers, respectively, we choose $\rho_{shared}=\rho_{task} = 0.95$ based on validation set performance.  

\noindent \textbf{Training}.
To train our MTL model, we use a traditional weighted-sum of task-specific losses:
\begin{equation}
\begin{aligned}
L_{MTL} = \sum_{i=1}^{T} w \times L_{i}
\end{aligned}
\end{equation}
where T is the set of tasks. As for task-specific losses, we use the pixel-wise cross-entropy for semantic segmentation and human-part segmentation, and the $L1$ loss for both depth estimation and surface normals estimation. Finally, we use the balanced-cross entropy for saliency detection. Similarly to MTLoRA, we use the same task-specific weights as \cite{MTI-NET}.

\subsection{Baselines}
The performance of FAAR was compared to the baselines used for MTLoRA \cite{mtlora}, similarly, we partition the baselines into models able to provide multiple task outputs with a single inference and other methods. We compare both efficiency and accuracy to other single-task and multi-task PEFT methods. We distinguish single-task PEFT baselines, obtained by running the method on each task respectively. First, we acquire a \textbf{Single Task} baseline by full fine-tuning a model, using the same pretrained model, for each task. \textbf{MTL} consists of training task-specific decoders from scratch while (1) having a pre-trained, frozen backbone (\textit{Tuning Decoders Only}), and (2) fine-tuning the backbone (\textit{Full Fine Tuning}). \textbf{Adapter} \cite{adapter} applies task-specific lightweight modules for each Transformer Layer. \textbf{Bitfit} \cite{bitfit} tunes only the biases, patch merging layers and patch projection layers. \textbf{VPT \cite{VPT}} performs prompt tuning either at the input level (VPT-shallow) or all transformer stages (VPT-deep). \textbf{Compacter} \cite{compacter} uses small shared adapter layers inside each Transformer block. \textbf{Compacter++} only updates MLP layers. \textbf{LoRA} \cite{lora-original} performs low-rank adaptation to attention layers. \textbf{Hyperformer} \cite{Hypernetwork} leverages a hypernetwork to generate adapter weights based on task embeddings.  \textbf{Polyhistor} \cite{polyhistor} introduces low-rank hypernetworks which learn to scale fine-tuning parameters inside transformer blocks. \textbf{MTLoRA} \cite{mtlora} performs LoRA-based adaptation and uses different LoRA modules to partition the pre-trained backbone into task-agnostic and task-specific parameters. \textbf{TADFormer} \cite{tadformer} alleviates the need for multi-task LoRA modules introduced by MTLoRA \cite{mtlora} by conditioning the LoRA modules on the task embedding via dynamic convolutions \cite{dynamic-filter}. \textbf{DiTASK} \cite{DiTASK} uses an alternative to traditional LoRA and leverages neural diffeomorphisms to adapt the singular values of each parameter matrix.

\subsection{Quantitative Analysis}
\cref{tab:multitask_pascal} compares the performance of FAAR to baselines, single-task comparable methods as well as MTL models on the Pascal-Context dataset \cite{pascal-context}. The number of trainable parameters is calculated on the whole model (\textit{encoder+decoders}). All methods presented use a \textit{Swin-Tiny} backbone pretrained on ImageNet-1k \cite{imagenet}. We observe that our method outperforms all other presented baselines. Similarly, in \cref{tab:multitask_nyud}, we report results obtained on  NYUD-v2 \cite{nyud}. Here, we compare our method to single-task and multi-task baselines, as well as best models from previous lora-based MTL PEFT strategies \cite{mtlora, tadformer}. Similarly as in \cref{tab:multitask_pascal}, we observe a performance gap between simple MTL and single-task baselines, the former performing worse. However, although no presented PEFT method manages to out-perform single-task training, FAAR outperforms MTL with full fine tuning while using fewer parameters. The gap between single-task and PEFT MTL performance can be explained by the smaller number of images in NYUD-v2 \cite{nyud}, making the adaptation optimization more difficult. As a result, we highlight in bold, in both tables, that our paper achieves better performance on both accuracy and efficiency metrics.

\subsection{Ablation Study}
The impact of our main innovations are reported in \cref{tab:new_ablation}. \textbf{1) FAAR} refers to the best configuration of our method as presented in \cref{sec:imp}. \textbf{2) w/o PDRS} refers to the removal of our rank shrinking method.  \textbf{3) w/o DoRA} refers to the change of DoRA adapters for LoRA adapters on top of our PDRS methods. Subsquently, \textbf{4) w/o TS-PD} refers to the ablation of our frequency-centric decoders on top of our DoRA adapted rank shrinking strategy. Finally, \textbf{5) w/o XT-Cons} refers to the sole removal of the cross-task operation inside our TS-PD.


\begin{table*}[!t]
\centering
\caption{Performance comparison on NYUD-v2 \cite{nyud} of LoRA-based MTL methods using \textit{Swin-Tiny} pretrained on ImageNet-1k as the backbone. We report $\Delta m$ as the relative improvement vs.\ single-task full fine-tuning.}
\label{tab:multitask_nyud}
{\setlength{\tabcolsep}{7pt}%
\renewcommand{\arraystretch}{1.0}%
\small
\begin{tabular}{c|ccc|c|c}
\hline
\noalign{\vskip 3pt}
\rule{0pt}{2.2ex}\textbf{Method} &
\textbf{SemSeg} \textit{(mIoU $\uparrow$)} &
\textbf{Depth} \textit{(rmse $\downarrow$)} &
\textbf{Normals} \textit{(rmse $\downarrow$)} &
$\Delta m$ (\%) &
\textbf{Trainable} \textbf{Parameters (M)}\\
\hline
Single Task & 42.65 & 0.60 & 22.83 & 0 & 84.00\\
MTL - Full Fine Tuning & 38.85 & 0.66 & 24.33 & -8.49 & 28.10 \\
\hline
MTLoRA \cite{mtlora} ($r{=}64$) & 39.85 & 0.66 & 29.53 & -15.30 & 9.70 \\
TADFormer \cite{tadformer} ($r{=}64$) & 40.85 & 0.64 & 27.48 & -10.42 & 8.90 \\
DiTASK \cite{DiTASK} & 41.13 & 0.65 & 27.25 & -10.45 & 3.06 \\
\textbf{FAAR} ($r_{init}{=}64$) & \textbf{41.27} & \textbf{0.63} & \textbf{26.35} & \textbf{-7.88} & \textbf{2.85} \\
\hline
\end{tabular}
}
\end{table*}

\begin{figure}[!t]
\centering
\includegraphics[width=\linewidth]{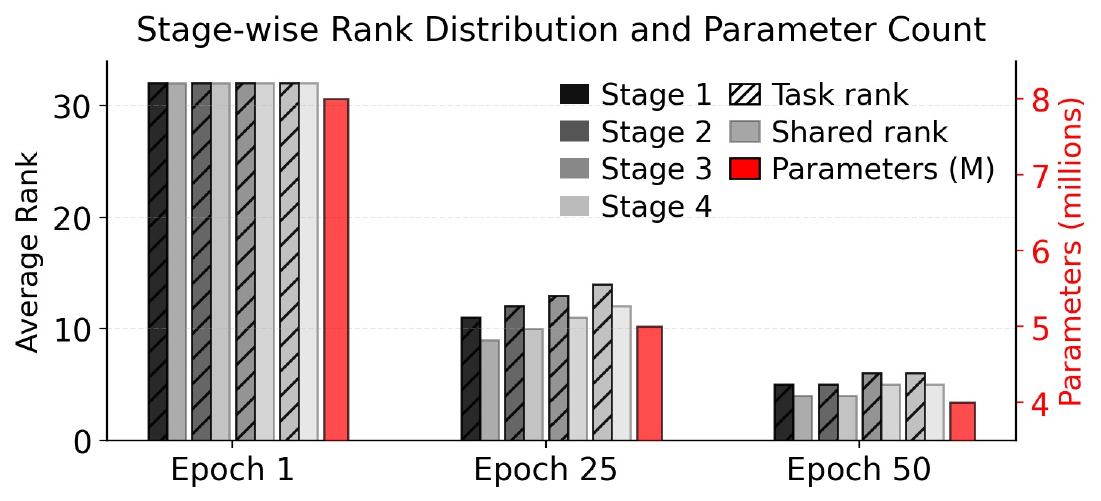}
\caption{Rank distribution across task-specific and shared layers for each transformer stage during early training.}
\label{rank-distrib}
\end{figure}

\begin{table}[t]
\centering

{\setlength{\tabcolsep}{1pt}%
\renewcommand{\arraystretch}{1.0}%
\footnotesize
\begin{tabular*}{\linewidth}{@{\extracolsep{\fill}} l c c c c c @{}}
\hline
\noalign{\vskip 2pt}
\rule{0pt}{2.1ex}\textbf{Model} &
\shortstack{\textbf{Sem.}$\uparrow$} &
\shortstack{\textbf{H.P}$\uparrow$} &
\shortstack{\textbf{Sal.}$\uparrow$} &
\shortstack{\textbf{Nor}$\downarrow$} &
\shortstack{$\Delta m$} \\
\hline
MTLoRA [2] (high) & 67.90 & 59.84 & 65.40 & 16.60 & +2.55 \\
MTLoRA [2] (low) & 65.57 & 57.93 & 63.48 & 18.54 & -2.56  \\
high w/ DoRA & 67.55 & 60.00 & 64.70 & 17.20 & +1.36 \\
low w/ DoRA & 65.52 & 58.06 & 63.65 & 18.43 & -2.31 \\
\hline
high + PDRS w/ LoRA      & 68.11 & 59.93 & 65.54 & 16.50 & +2.83 \\
high + PDRS w/ DoRA (1)      & 71.35 & 61.02 & 65.92 & 16.42 & +4.92 \\
\hline
high + TS-PD             & 69.88 & 60.26 & 65.90 & 16.47 & +3.83 \\
high + TS-PD + XT-Cons (2)   & 70.73 & 60.95 & 65.92 & 16.40 & +4.63 \\
\hline
\textbf{FAAR} (1+2) & \textbf{72.02} & \textbf{61.25} & \textbf{66.11} & \textbf{16.35} & +5.28 \\
\hline
\end{tabular*}
}
\caption{Additive Component Analysis on PASCAL}
\label{tab:new_ablation}
\end{table}

\mycomment{
\begin{table}[t]
\centering
\caption{Hierarchical ablation on PASCAL~\cite{pascal-context}.}
\label{tab:ablation}
{\setlength{\tabcolsep}{3.2pt}%
\renewcommand{\arraystretch}{1.06}%
\footnotesize
\begin{tabular*}{\linewidth}{@{\extracolsep{\fill}} l c c c c c @{}}
\hline
\noalign{\vskip 2pt}
\rule{0pt}{2.1ex}\textbf{Model} &
\shortstack{\textbf{SemSeg}\\\textit{(mIoU $\uparrow$)}} &
\shortstack{\textbf{HumanParts}\\\textit{(mIoU $\uparrow$)}} &
\shortstack{\textbf{Saliency}\\\textit{(mIoU $\uparrow$)}} &
\shortstack{\textbf{Normals}\\\textit{(rmse $\downarrow$)}} &
\shortstack{$\Delta m$\\(\%)} \\
\hline
\textbf{FAAR}        & \textbf{72.02} & \textbf{61.25} & \textbf{66.11} & \textbf{16.35} & \textbf{+5.28} \\
w/o PDRS             & 69.53 & 60.10 & 65.21 & 16.58 & +3.20 \\
w/o DoRA             & 70.78 & 60.35 & 65.34 & 16.45 & +4.00 \\
w/o TS--PD           & 71.24 & 60.80 & 65.89 & 16.45 & +4.58 \\
w/o XT--Cons         & 71.55 & 60.98 & 66.01 & 16.41 & +4.87 \\
\hline
\end{tabular*}
}
\end{table}
}

\mycomment{
\begin{figure}[!t]
\centering
\includegraphics[width=0.9\linewidth]{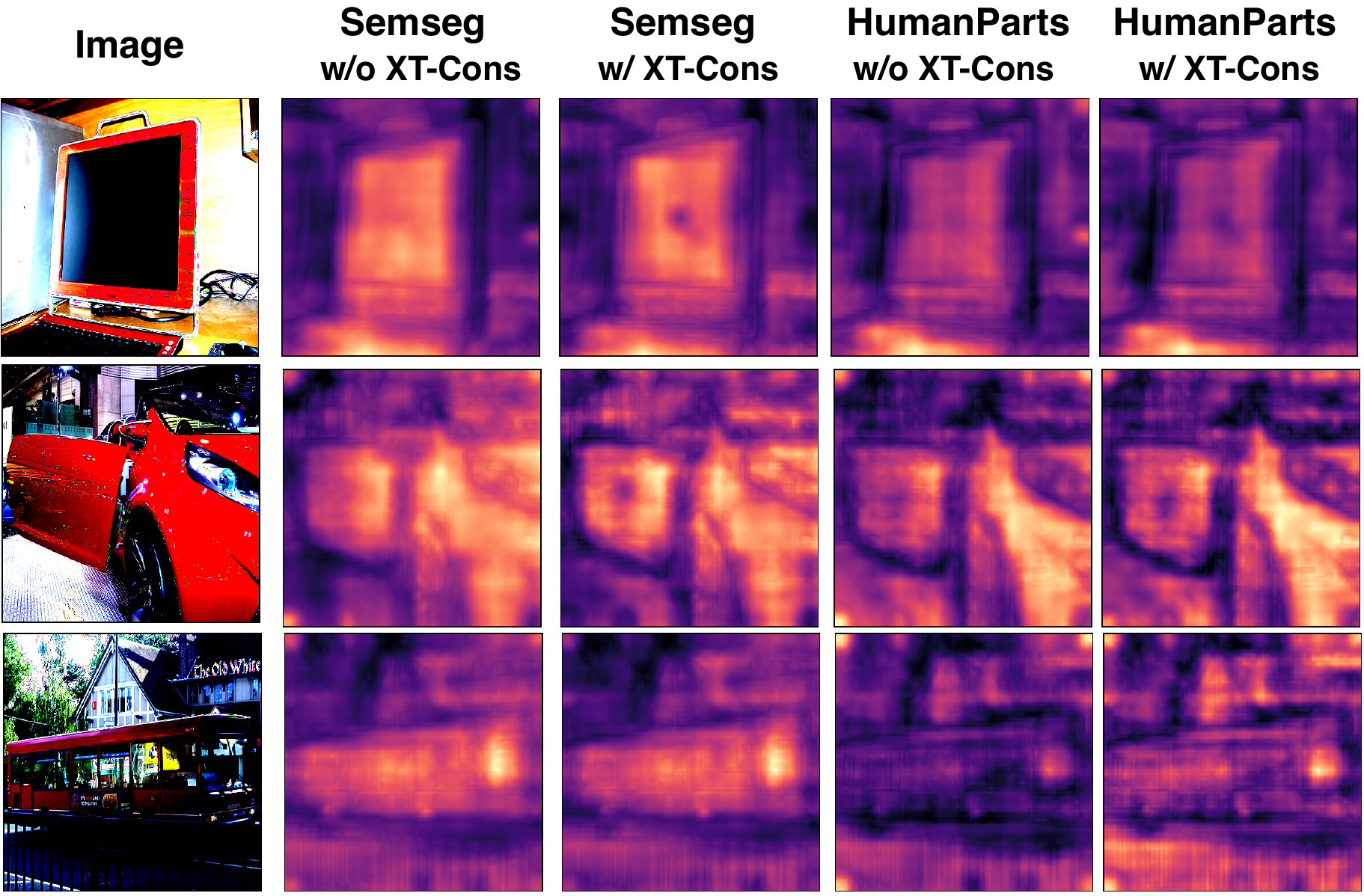}
\caption{Visualization of the effect of our TS-PD decoder. Notice how our XT-Cons enforces edge alignment. For simplicity, only segmentation and human part segmentation features are reported.}
\label{XT-Cons-fig}
\end{figure}}

\textbf{Effectiveness of PDRS}. The reduction of rank-1 components in our low-rank adaptation is a key factor to the drastic drop in number of parameters brought by FAAR. We detail this process in \cref{rank-distrib}. Starting from a FAAR configuration with $r_{init}=32$, this figure represents the evolution of ranks throughout the first 50 epochs, and shows the large reduction, of 4 millions, in parameters caused by the reduction in rank. It can be seen that, there is a tendency for task-specific low-rank adaptors to exhibit bigger ranks, similarly there is also a tendency for deep layers (stage 3 and 4) to have bigger ranks than shallow layers. This is explained by the fact that decision making layers are both task-specific layers and deep layers which exhibit finely detailed information, crucial for dense tasks. 

\textbf{DoRA vs LoRA}. Our method leverages DoRA \cite{DoRA} due to its stronger performance on low ranks values. \cref{tab:new_ablation} shows that on two instances of MTLoRA \cite{mtlora}. One with a high rank (r = 64) and a low rank (r = 4). If we swap LoRA for DoRA, we observe the performance for high rank is degraded whereas the performance for low rank is somewhat improved. FAAR, which starts from $r_{init}=64$ to quickly converge to $r_{global} \approx 5$, gets major improvement as DoRA works particularly well in combination with our PDRS.

\textbf{Effectiveness of TS-PD and XT-Cons}. TS-PD produces an overall improvement of $\Delta_{m}=+3.83$ (\cref{tab:new_ablation}). The addition of our XT-Cons method brings an improvement of $\Delta_{m}=+4.63$. Proving, in the line of other works \cite{MT-CP, PAD-NET, MTI-NET} that leveraging other task representations to maintain geometric consistencies across tasks is a useful aspect of MTL models. Our XT-Cons is therefore a cheap and effective way to improve the performance of all the tasks. 

\textbf{Effect of initial rank budget and coverage criterion}. The parameter $r_{init}$ defines the initial rank of all layers at the start of training. We find that the initial rank needs to be high enough to allow PDRS to prune sufficient rank-1 directions. However, we notice no difference in the overall performance for $r_{init} \in {64,32,16}$ suggesting that these starting values provide enough search space for rank directions to be learned. Also, the coverage criterion $\rho_{shared}$ and $\rho_{task}$ define the strength of the rank shrinking decision (see \cref{rho_eq}). Therefore, a stronger value will lead to slower decrease whereas a lower value will lead to steep and fast reduction. Our chosen values are reported in \cref{sec:implementation_details} and are based on validation set performance.


\section{Conclusion}
We have introduced a novel, low-rank MTL PEFT method, named FAAR, that efficiently adapts a frozen, pre-trained backbone for multiple downstream dense visual tasks. To allow each of the tasks to have a flexible and powerful adaptation, we propose PDRS, a dynamic method which drastically saves on computation power by reducing the rank of each layer in a backbone during training, while maximizing the MTL performance. Moreover, to capture fine-grained details, relevant to dense visual tasks, we propose TS-PD, a FFT-centric approach that uses strong geometric cues in a task-specific manner. To a further extent, TS-PD uses these cues to perform cross-task consistency, to align the dense visual tasks along the geometric information of the input image. Experiments on two representative MTL benchmarks show that FAAR outperforms previous works from both an accuracy and efficiency perspective. Future research includes leveraging FAAR as a way to increase the number of tasks as part of MTL models due to the induced, reduced cost of backbone fine-tuning.\\
\textbf{Acknowledgment.} The authors would like to thank Prof. Tomasz Radzik
for helpful discussions. Computing resources provided by
King’s Computational Research, Engineering and Technology Environment (CREATE). M. Fontana is supported by the NMES studentship funding; M. Shi is supported by Fundamental Research Funds for the Central Universities.

{
    \small
    \bibliographystyle{ieeenat_fullname}
    \bibliography{main}
}


\end{document}